\pdfoutput=1

\documentclass[11pt,a4paper]{article}

\usepackage{eamt26}
\usepackage{times}
\usepackage{url}
\usepackage{latexsym}

\usepackage[small,bf]{caption} 
\usepackage[T1]{fontenc}
\usepackage[utf8]{inputenc}
\usepackage{microtype}
\usepackage{inconsolata}
\usepackage{graphicx}
\usepackage{enumitem}
\usepackage{booktabs}
\usepackage{multirow}

\usepackage{array}
\usepackage{xcolor}

\usepackage[avoid-all]{widows-and-orphans}

\usepackage{xparse}
\NewDocumentCommand{\citep}{O{} O{} m}{\cite{#3}}
\NewDocumentCommand{\citet}{O{} O{} m}{\cite{#3}}

\newcommand{\citeauthor}[1]{\cite{#1}}
\newcommand{\citeyear}[1]{n.d.}
\usepackage[hidelinks]{hyperref}
\makeatletter
\renewcommand\@biblabel[1]{}
\makeatother
\newenvironment{compactitemize}
  {\begin{itemize}[itemsep=0pt, parsep=0pt, topsep=0pt]}
  {\end{itemize}}

\title{Translation Analytics for Freelancers II:\\Benchmarking Local LLMs for Confidential Translation Workflows}

\author{Yuri Balashov\thanks{\ \hspace{1pt} Equal contribution.\ Corresponding author.} , Rex VanHorn\footnotemark[1] , Mingxi Xu, Austin Downes\\
  University of Georgia, Athens, Georgia, USA\\
  {\tt \{yuri,rex.vanhorn,mingxi.xu,austin.downes\}@uga.edu}}

\begin{document}
\raggedbottom
\maketitle
\enlargethispage{8pt}

\begin{abstract}
Building on our previous work, this paper develops practical, low-barrier methods for freelance translators and smaller language service providers to evaluate translation technologies using rigorous yet accessible analytic methods.\ Here we address a high-stakes, specialized need: offline translation for confidentiality-sensitive domains in which privacy constraints preclude the use of cloud-based engines and commercial LLMs.\ We expand the Reeve Foundation Trilingual Corpus (RFTC) used in our previous work into a multilingual corpus (RFMC) by adding sentence-aligned German and Simplified Chinese reference translations. We then benchmark several locally runnable language models (via Ollama) across four language directions on 1000+ sentences selected from this corpus. We use consistent single-prompt calls without fine-tuning or domain adaptation, comparing local LLM outputs against commercial NMTs (DeepL, Baidu), a frontier LLM (GPT-5.2), and professional-grade local NMT systems (OPUS-CAT, NeuralDesktop, Promt). Automatic evaluation is conducted with MATEO. Results reveal substantial variation in local LLM performance across language directions and model sizes. The best local LLMs match or surpass local NMT systems and a frontier LLM, though they remain behind top commercial NMTs. These findings underscore the viability of carefully selected local LLM translation for privacy-constrained professionals and inform future research on model scaling and multilingual capability.
\end{abstract}

\section{Introduction}
\label{sec:introduction}

Language Service Providers (LSPs), large and small, are increasingly using large language models (LLMs) in their localization workflows. Freelance translators do it too in sporadic and ad hoc ways \citep{penet_teaching_2025}. In an earlier paper \mbox{\citep{balashov_translation_2025}}, we demonstrated the value of simple analytic tools---automatic metrics, structured human assessment, and lightweight statistical analysis---that were historically confined to MT research and large-scale industrial QA but are increasingly accessible to individual translators and small LSPs. Using the Christopher \& Dana Reeve Foundation Trilingual Corpus (RFTC) derived from a real medical-domain translation project, we showed how freelancers can compute and interpret automatic evaluation metric scores (BLEU, chrF, TER, COMET) with minimal infrastructure and can correlate those scores with structured human judgments to determine which metrics track translator-perceived quality in a specific setting. A key takeaway was pragmatic: evaluation need not be ``big tech only.'' Even modest, sample-based analyses can be informative and statistically stable for system comparison in a real workflow. Our findings emphasize the importance of proactive engagement with modern technologies and analytic methods to not only adapt but thrive in the rapidly evolving professional environment.

In our second paper, we apply this framework to a niche area of the translation services market. We focus on the base-level translation performance of smaller language models that can be installed and run locally by users with modest resources and no special programming skills. This is important for several reasons outlined below.

\subsection{Practical Considerations: Why Offline Translation May Be Required}
\label{sec:practical}

In parallel with the spread of cloud MT and frontier LLMs, translation practice has also seen a countervailing demand: strict data-control workflows \citep{berger_machine_2024,dogru_data_2024,lyu_paradigm_2024,sandrini_beyond_2025}. Certain projects in defense, security, finance, and patent-related work require offline processing or air-gapped environments. In such scenarios, API-based commercial systems are not suitable, regardless of quality. This creates a concrete need for translators to evaluate and deploy locally runnable technologies---either local NMT toolchains\footnote{For example, OPUS-CAT \citep{nieminen_opus-cat_2021} built around OPUS-MT models \citep{tiedemann_opus-mt_2020} and designed for offline professional use. Other locally installable NMT systems include NeuralDesktop and Promt (introduced as baselines in Section \ref{sec:design} below).} or newer open-weight LLMs that can be executed locally through end-user applications.

Local LLM execution has rapidly become more feasible for non-programmers due to desktop inference tools and model packaging ecosystems. The remaining question for freelancers is the empirical one we address below: whether base translation quality from small-to-mid-size local LLMs is good enough to support confidential professional workflows.

\subsection{Theoretical Considerations: Scaling Laws and Interpretability}
\label{sec:theoretical}

From a more theoretical perspective, studying the translation performance of smaller language models of different sizes, such as llama3.1-8b, gpt-oss-20b, or gemma3-27b, could cast additional light on ``scaling laws.'' The earlier literature on neural language modeling shows robust scaling-law behavior for model performance as a function of parameters, data, and compute \citep{kaplan_scaling_2020,hoffmann_training_2022}. Translation provides a concrete capability domain where scaling may or may not behave monotonically across languages and writing systems \citep{ghorbani_scaling_2021,alves_tower_2024,rei_tower_2024,zhu_multilingual_2024,richburg_how_2024}. Our multilingual corpus and a broad selection of local LLMs of different sizes and provenance give us an opportunity to study this behavior in a practical setting.

Additionally, compared with many proprietary systems, open-weight releases often provide more transparent model cards, training descriptions, and language coverage \citep[e.g.,][]{google_deepmind_gemma_2025,qwen_team_qwen25_2025,qwen_team_qwen3_2025,dang_aya_2024,meta_llama_2025}. Observed translation differences can therefore be discussed in relation to architecture and multilingual training choices with fewer unknowns.

\subsection{Related Work}
\label{sec:related}

The application of LLMs to translation and other multilingual tasks began essentially as soon as LLMs themselves emerged. Brown et al.\ \shortcite{brown_language_2020} demonstrated that GPT-3 could perform zero- and few-shot translation, and subsequent work rapidly explored the boundaries of this capability. Major evaluations by Hendy et al.\ \shortcite{hendy_how_2023}, Jiao et al.\ \shortcite{jiao_is_2023}, and Vilar et al.\ \shortcite{vilar_prompting_2023} suggested that frontier LLMs could achieve translation quality competitive with dedicated NMT systems for high-resource language pairs, though gaps remain for low-resource directions and specialized domains. A comprehensive recent survey by Ataman et al. \shortcite{ataman_machine_2025} traces the trajectory from early neural MT architectures through the current LLM era, highlighting both the promise and the open problems including domain adaptation, evaluation methodology, and the role of parallel versus monolingual training data. Lyu et al.\ \shortcite{lyu_paradigm_2024} argue that a paradigm shift is underway in which LLMs will increasingly subsume traditional NMT, particularly for document-level and stylized translation. Prompting, few-shot, and iterative-refinement strategies have continued to push LLM translation quality upward \citep{zhang_prompting_2023,peng_towards_2023,garcia_unreasonable_2023,vilar_prompting_2023,he_exploring_2024,briva-iglesias_large_2024,briakou_translating_2024,chen_iterative_2024,berger_prompting_2024,aldosari_assessing_2025,feng_tear_2025,rajaee2026}. Alongside general-purpose families, dedicated translation LLMs such as Tower \citep{alves_tower_2024,rei_tower_2024} and, more recently, SalamandraTA \citep{gonzalezagirre2025salamandratechnicalreport} and Tower+ \citep{rei2025towerbridginggeneralitytranslation} have continued to improve open-weight translation quality at modest scale.

The more recent rollout and rapidly increasing accessibility of local language models, facilitated by inference platforms such as Ollama \citep{ollama_ollama_2024} and LM Studio \citep{lm_studio_lm_2024}, have led MT researchers and translation scholars to explore the translation capabilities of models that can run entirely on consumer hardware. Cui et al.\ \shortcite{cui_multilingual_2025} systematically evaluate open LLMs with fewer than ten billion parameters on multilingual MT tasks, finding that models like Gemma2-9B exhibit impressive multilingual translation capabilities. They introduce the GemmaX2-28 model, which achieves competitive performance with Google Translate and GPT-4-turbo across 28 languages. Sandrini \shortcite{sandrini_beyond_2025} investigates the feasibility of locally deployable, free language models as alternatives to proprietary cloud-based solutions from the perspective of practicing translators, evaluating three open-source models (llama3-8b, mixtral-8x7b, and gemma2-27b) with three toolkits (GPT4ALL, Llamafile, and Ollama) to generate translations in tourism/marketing and legal domains from Italian to German. In Sandrini's setup, MATEO was used to score those translations against outputs from ChatGPT-4.0-mini and Gemini-1.5-flash, which served as reference translations.

Our work falls in this category but differs from both \citet{cui_multilingual_2025} and \citet{sandrini_beyond_2025} in important respects. Unlike the former, our main audience comprises users, not developers, of translation technologies. Additionally, while ~Sandrini's \shortcite{sandrini_beyond_2025} user study is pioneering in its category, we take it to be subject to several limitations: the sample size is modest (fewer than twenty-five sentences); inference is performed on a CPU, typically relying on lower-precision quantization and suboptimal backends, which may introduce greater output variance than GPU-based higher-precision inference; and LLM outputs, rather than professional human translations, serve as reference translations.

We study a broader range of more recent local LLMs using a medical corpus originating from a real translation project. While our source language is English, our target languages (DE, RU, JA, ZH) comprise a typologically diverse set with different writing systems, thus allowing us to investigate interesting cross-linguistic phenomena. While we only use Ollama for all our translation calls, our sample is much larger (over a thousand sentences) and the translation behavior is more stable due to our use of a consumer-grade GPU. The larger sample makes our results statistically significant.

\subsection{Our Contributions}
\label{sec:contributions}

This paper contributes:

\begin{compactitemize}
    \item An expansion of the Reeve Foundation Trilingual Corpus (RFTC) to German and Simplified Chinese. The resulting Reeve Foundation Multilingual Corpus (RFMC) now includes approximately 3,500 source sentences in the medical domain aligned with their recent professional translations to four typologically different languages: EN$\to$\{DE, RU, JA, ZH\}. With the client's permission, we make the RFMC available for noncommercial/academic use.
    \item A freelancer-oriented benchmark of 9 locally runnable LLMs against commercial NMT, a frontier LLM, and local NMT baselines, under ``base translation only'' constraints.
    \item A meta-analysis of prompting and decoding strategies.
\end{compactitemize}

What makes our approach distinctive is the combination of high-quality data from a recent professional translation project, systematic leverage of accessible small LLMs, and the translation experience of some team members, including an ATA-certified professional translator, which informs our evaluation methodology and interpretation of results. We frame this work explicitly as a \emph{user}-oriented benchmark, written by users of translation technology, for tech-savvy freelance translators and smaller LSPs who have only modest resources and no or little programming skills.\ The methods and code are intended to be accessible to this audience.

The remainder of the paper presents corpus expansion (Section~\ref{sec:corpus}), experimental design (Section~\ref{sec:design}), translation generation (Section~\ref{sec:generation}), automatic evaluation (Section~\ref{sec:evaluation}), linguistic observations (Section~\ref{sec:discussion}), and implications for practice-, scaling-, and cost-oriented research (Section~\ref{sec:conclusion}).

\section{Corpus Expansion}
\label{sec:corpus}

The Reeve Foundation Trilingual Corpus (RFTC), introduced in \citet{balashov_translation_2025}, pairs English source sentences from the International Edition of the Christopher \& Dana Reeve Foundation's Paralysis Resource Guide (PRG) with their recent professional translations to Russian and Japanese. To enable a broader, multilingual study, we expanded the RFTC by adding two additional target languages: German (DE) and Simplified Chinese (Mainland China; ZH).

The expansion required aligning the existing professional translations of PRG International in German and Chinese, published by the Reeve Foundation as PDF documents, with the existing trilingual segments (EN--RU--JA) already contained in the RFTC. This entailed two main steps.\ First, we extracted text from the PDF files in DE and ZH, performing cleanup and downsizing analogous to those described in Section~2 of \cite{balashov_translation_2025}. Second, we undertook corpus curation to address a version mismatch. The existing translations of the PRG International to RU and JA, already contained in the RFTC, derive from the most recent English edition (2023). However, the official translations of PRG International to DE and ZH were based on the earlier English edition (2022). As a preliminary step, we identified the segments common to both editions (approximately 70\% of the corpus) and retained the existing DE and ZH translations for those segments. For the remaining segments (approximately 30\%), which had been added or modified in the 2023 edition, we updated the translations using expert knowledge of DE and ZH.

The resulting Christopher \& Dana Reeve Foundation Multilingual Corpus (RFMC) presents approximately 3,500 source sentences (EN) aligned with their original professional translations to RU and JA, as well as their partially curated reference translations to DE and ZH. The RFMC is made available on GitHub for noncommercial/academic use.\footnote{\url{https://github.com/YuriBalashov/reeve-mftc}}

\section{Experimental Design}
\label{sec:design}

We selected a continuous portion of RFMC---the entire cleaned and curated Chapter~1 of PRG~International-2023 containing over a thousand sentences---for our experiments to explore the translation performance of 9~local language models available on Ollama \citep{ollama_ollama_2024}:

\begin{compactitemize}
    \item aya-expanse-32b \citep{dang_aya_2024}
    \item deepseek-r1-32b \citep{deepseek-ai_deepseek-r1_2025}
    \item gemma3-27b \citep{google_deepmind_gemma_2025}
    \item gpt-oss-20b \citep{openai_introducing_2025}
    \item llama3.1-8b-instruct-fp16 \citep{llama_team_llama3_2024}
    \item mistral-small3.2-24b \citep{mistral_ai_mistral_2025}
    \item qwen2.5-32b \citep{qwen_team_qwen25_2025}
    \item qwen3-32b \citep{qwen_team_qwen3_2025}
    \item translategemma-27b \citep{finkelstein_translategemma_2026}
\end{compactitemize}

\noindent and compared their performance with three important baselines:

\begin{compactitemize}
    \item The best-performing commercial NMT system: DeepL for EN--DE, EN--RU, and EN--JA; Baidu for EN--ZH\footnote{\url{https://www.deepl.com/en/translator}; \url{https://www.baidu.com}}
    \item The best-performing frontier commercial LLM: GPT-5.2 \citep{openai_introducing_2025-1}
    \item The best-performing generally available local NMT system: Promt-23 for EN--RU, EN--JA, and EN--ZH; NeuralDesktop for EN--DE\footnote{\url{https://www.promt.com}; \url{https://neuraldesktop.com}}
    \end{compactitemize}

\noindent This section describes our model and prompt selection process.

\subsection{Model Selection}
\label{sec:model-selection}

Our model selection was guided by several practical constraints reflecting the resources typically available to tech-savvy freelance translators and smaller LSPs. We sought open-weight models that (i)~could run locally via Ollama on a consumer-grade GPU, (ii)~covered a range of parameter counts from 8B to 32B to allow investigation of scaling effects, (iii)~represented diverse model families and provenance (open-weight releases from Google, Meta, Mistral, Alibaba, Cohere, DeepSeek, and OpenAI), (iv)~routinely or historically performed well on translation benchmarks, and (v)~included both general-purpose and translation-specialized architectures.

Our inference hardware consisted of NVIDIA GeForce RTX 3090 GPUs (24~GB VRAM), a consumer-grade card widely available on the secondary market at the time of writing. Each translation run was executed on a single GPU, reflecting the setup a typical freelancer or a small LSP might reasonably assemble. Although our workstation housed multiple cards, they were used only to parallelize independent experiments; no model was distributed across GPUs, and all reported results reflect single-card performance.

The inference engine was Ollama v0.14.2, running on Ubuntu 24.04. Ollama provides a straightforward command-line interface for downloading, managing, and serving open-weight models, requiring no programming expertise beyond basic terminal familiarity.\ We note that comparable results would be expected from other llama.cpp-based inference frontends, such as LM Studio, which offers a graphical interface and may be more approachable for users less comfortable with the command line.

Models at or above 24B parameters (aya-expanse-32b, deepseek-r1-32b, gemma3-27b, mistral-small3.2-24b, qwen2.5-32b, qwen3-32b, translategemma-27b) were served in Ollama's default 4-bit quantization (Q4\_K\_M), balancing inference speed and memory footprint against output quality. The smaller models (gpt-oss-20b, llama3.1-8b-instruct) fit comfortably within the 24 GB VRAM budget without quantization: gpt-oss-20b was run at its default precision, while llama3.1-8b-instruct was run at full fp16 to explore whether preserving numerical fidelity at a smaller scale could compensate for fewer parameters. As a general principle, we selected the lowest quantization level (i.e., highest precision) that could reasonably fit into 24 GB of VRAM. All models were loaded entirely into GPU memory for each run. Because our translation pipeline processes each sentence as an independent API call with no carried context (Section~\ref{sec:local-llm-outputs}), the effective context window per request was short, and VRAM was not a binding constraint for any model in our selection.

We note that while by industry standards, these are modest resources, most freelancers, even technically-inclined ones, may not have immediate access to comparable hardware. However, the landscape is changing rapidly, with GPUs and bundled inference engines increasingly packaged into new PC and Mac consumer offerings, and cloud GPU rental services becoming more affordable. Further details on hardware specifications and runtime configurations are provided in Appendix~\ref{app:inference}.

\subsection{Prompt Selection}
\label{sec:prompt-selection}

Recent work on using LLMs for translation suggests that smaller models' outputs tend to be more sensitive to prompting details than the outputs from larger models \citep{zhang_prompting_2023,peng_towards_2023,zhu_multilingual_2024,aldosari_assessing_2025}. Since our explicit goal is to study the translation performance of smaller language models, we began with pilot experiments on a smaller set of 118 test sentences from other parts of PRG International not overlapping our main source document. These sentences were manually selected to cover linguistic phenomena representative of the entire corpus: sentences containing URLs, named entities (organization, program, and product names; personal proper names), technical terms and acronyms; and longer clauses.

The models were then asked to translate these 118 sentences to a single language from our target set (DE) with 11 different candidate prompts. Those candidate prompts were compiled as follows: our 8 selected local language models and 3 frontier LLMs (Claude Opus 4.5, GPT-5.2, and Gemini Pro 3) were ``meta-prompted'' to generate their own preferred prompt for translation:

\begin{quote}
\small
We need a system prompt for a translation task. The requirements are:\\
- Source language: English\\
- Target language: German\\
- Audience: expert/academic\\
- Output: translation only, no explanations, annotations, or transliterations\\
Write the prompt you would want to receive if you were performing this task. Output only the prompt text itself, nothing else.
\end{quote}

Candidate prompts were elicited from each model: local models via the Ollama API used later for translation, frontier LLMs via context-free manual chat sessions.\ Two prompt suggestions, from llama3.1-8b-instruct-fp16 and aya-expanse-32b, were judged inadequate and discarded.\ The prompt from llama3.1-8b-instruct was discarded because it produced an improperly formatted response that would not have functioned as a system prompt, and we chose not to manually correct it to avoid introducing experimenter bias.

Interestingly, aya-expanse-32b attempted to create its preferred translation prompt in German instead of English.\ This intent is consistent with recent studies that explicitly prompt LLMs in languages other than English, sometimes resulting in improved performance on multilingual tasks \citep{nguyen_democratizing_2024,mondshine_beyond_2025,gupta_how_2025}. We decided to let our models explore this opportunity and included a German version of a ``standard'' prompt from \citet{balashov_translation_2025} alongside its English version.

Each of the 11 candidate prompts (p1--p11; see Appendix~\ref{app:prompts}) compiled in this way was then used to query each of our eight models to translate 118 held-out test sentences from EN to DE. Thus, every model generated a translation from every candidate prompt. The COMET-22 scores for the resulting 88 translation outputs (Table~\ref{tab:prompt-pilot}) do not, by themselves, mark a clear winner among and do not appear to correlate with the ``native'' prompt choice: in many cases the prompt suggested by model~$X$ failed to maximize the COMET scores for~$X$'s own output.

Accordingly, we decided to look for a prompt that was best ``on average.''\ Two methods were used for this purpose: (i)~maximizing the sum of $z$-scored COMET improvements over the prompt mean across models; and (ii)~rank aggregation across the models. Both methods selected p5 as ``the best prompt on average'':

\begin{quote}
\small
\textbf{p5:} Translate the following English text into \{target\_lang\}. The target audience is academic/expert. Provide *only* the \{target\_lang\} translation, without any explanations, annotations, or transliterations.
\end{quote}

During this pilot phase (mid-January 2026), TranslateGemma became available \citep{finkelstein_translategemma_2026}. We decided to add it to our 8 initial LLMs but used a different prompt (p0), which was explicitly recommended by its developers:

\begin{quote}
\small
\textbf{p0:} You are a professional \{source\_lang\} (\{src\_lang\_code\}) to \{target\_lang\} (\{tgt\_lang\_code\}) translator.\ Your goal is to accurately convey the meaning and nuances of the original \{source\_lang\} text while adhering to \{target\_lang\} grammar, vocabulary, and cultural sensitivities. Produce only the \{target\_lang\} translation, without any additional explanations or commentary. Please translate the following \{source\_lang\} text into \{target\_lang\}: \textbackslash n \textbackslash n \textbackslash n \{TEXT\}\
\end{quote}

\noexpand We used p0 only for TranslateGemma.\ Although this introduces an additional variable, TranslateGemma was the sole translation-specialized model in our panel, and it was trained with p0 (\emph{ibid.}, p.\ 5).\ We judged that respecting the developer-recommended prompt for a single model gave a fairer picture of its base translation capability than forcing a generic prompt onto it.

\subsection{Temperature Settings}
\label{sec:temperature}

We conducted temperature sensitivity experiments for a subset of our models across the range $T=0.0$ to $T=1.0$ in increments of 0.05. Consistent with prior findings on LLM-based translation \citep{peng_towards_2023,balashov_translation_2025}, translation quality as measured by COMET did not vary substantially across temperature settings. However, analysis of pairwise Levenshtein ratios across temperature conditions revealed a clear pattern: outputs generated at $T=0.0$ were maximally similar to outputs at all other temperatures, effectively serving as the centroid of the output distribution. As temperature increased, outputs diverged not only from the low-temperature baseline but increasingly from one another, with the lowest pairwise similarity observed between adjacent high-temperature settings (e.g., $T=0.95$ vs.\ $T=1.0$). These results suggest that for translation, temperature primarily affects output consistency rather than quality. We therefore set $T=0.0$ for our main experiments with ``non-reasoning'' models, to maximize reproducibility. A sample Levenshtein ratio matrix for the temperature sweep is provided in Appendix~\ref{app:temperature} and Figure~\ref{fig:levenshtein_ratio}.

However, we found that running three ``reasoning'' models with zero decoding temperature caused them to ``loop'' and time out, which is consistent with other recent studies \citep{pipis2025waitwaitwaitreasoning}. To minimize the negative impact of infinite ``reasoning loops'' we decided to use $T=1.0$ for gpt-oss-20b and $T=0.6$ for qwen3-32b (model card overrides), and $T=0.8$ for deepseek-r1-32b (Ollama default). We provide additional details on this in Section~\ref{sec:local-llm-outputs} and Appendix~\ref{app:failures}.

\subsection{Target Language Specification}
\label{sec:target-lang}

Prompting language models for translation requires explicit specification of a target language.\ In our case, the prompt recommendation for TranslateGemma also required specifying an ISO code for the target language. This was straightforward for German (DE), Russian (RU), and Japanese (JA). The initial options for Chinese included `Simplified Chinese', `Mandarin', `Mandarin Chinese', `Chinese (Mandarin)', `Simplified Chinese (Mandarin)', and `Simplified Chinese (Mainland China)'. Upon consultation with a native Chinese speaker familiar with the subject matter and with ChatGPT-5.2, the set of options was reduced to `Simplified Chinese (Mainland China)', `Simplified Chinese', and `Simplified Chinese (Mandarin)'. A quick translation of 24 English sentences selected from our corpus with three versions of the p5 prompt incorporating these options did not result in significant COMET score differences for any of our models. We chose to use `Simplified Chinese (Mainland China)' and `zh-CN' in our main experiments.

\section{Generation of Translation Outputs}
\label{sec:generation}

As noted earlier (Section~\ref{sec:design}), we selected a continuous part of RFMC---the entire curated Chapter~1 of PRG International-2023 containing 1,143 sentences---for our main experiments.

\subsection{Translation Outputs from Local Language Models}
\label{sec:local-llm-outputs}

We generated translations of the selected document to \{DE, RU, JA, ZH\}, using p5 (our ``best prompt an average,'' Section~\ref{sec:prompt-selection}) to interact with all the selected models except TranslateGemma, for which we used p0, as recommended by its developers. Outputs were stored in a normalized, line-aligned format to support automated scoring and later manual review.\footnote{Code available at \url{https://github.com/asdownes/translation-analytics-llm-benchmark}}

Each source sentence was translated through an independent API call to the local Ollama server, with the selected prompt (Section~\ref{sec:prompt-selection}) provided as the system message and the source sentence as the user message. No conversational context or translation memory was carried between calls. This sentence-by-sentence approach, while less sophisticated than document-level translation methods that leverage surrounding context \citep{hu_source-primed_2025}, eliminates confounding factors related to context window management and ensures straightforward reproducibility. It also mirrors the segment-by-segment workflow common in CAT tool environments, where MT suggestions are typically generated for individual translation segments. We note that providing additional context---for example, by constructing a lightweight translation memory from previously translated segments---could plausibly improve output quality for local LLMs, but investigating such strategies is outside the scope of the present study. Our results should therefore be understood as a lower bound on achievable translation quality with these models.

As already noted, six ``non-reasoning'' models were run at $T=0.0$, consistent with the findings reported in Section~\ref{sec:temperature}. Output post-processing was minimal: where a model returned extraneous formatting (e.g., markdown fences, list prefixes, or wrapping quotation marks), the output was automatically stripped to the translation content only, though such artifacts were infrequent with the models in our final selection.

The three reasoning-oriented models in our set---gpt-oss-20b, deepseek-r1-32b, and qwen3-32b---were the only models susceptible to generation failures due to their internal chain-of-reasoning processing. These failures occurred when the model entered repetitive reasoning loops, consuming the token budget without producing a final translation. gpt-oss-20b was by far the most affected, producing 99 failed lines across the four language directions when the decoding temperature was set to zero. Changing the temperature to 1.0 dramatically reduced the number of failed calls to 7, but also appears to have a negative effect on the output quality. Appendix~\ref{app:failures} provides further details.

All remaining models completed the full corpus without failure with $T=0.0$. Wall-clock translation times per language direction on a single RTX 3090 ranged from roughly 19–40 minutes for non-reasoning models to 2.9–5.5 hours for reasoning-oriented models (Table~\ref{tab:wallclock}, Appendix~\ref{app:inference}); a fuller per-model runtime breakdown will accompany the code release.

\subsection{Additional Translation Outputs}
\label{sec:additional-outputs}

For each of our translation directions (EN$\to$\{DE, RU, JA, ZH\}), we generated additional outputs from generally-available commercial NMT systems (Google Translate, DeepL, Baidu), a frontier LLM (GPT-5.2), and generally-available local NMT systems (Promt, NeuralDesktop, OPUS-CAT). These systems represent the spectrum of tools currently accessible to professional translators: cloud-based engines offering state-of-the-art quality, API-accessible frontier LLMs, and offline NMT solutions designed for confidential work. See Section~\ref{sec:design} for references to these systems.

As detailed below (Section~\ref{sec:evaluation}), based on the document-level COMET-22 scores of our outputs, we selected three systems for each language pair: the best-performing commercial NMT system, the best-performing frontier LLM, and the best-performing generally available local NMT system. These served as important baselines for our evaluations.

\section{Automatic Evaluation}
\label{sec:evaluation}

\subsection{Metrics and Tools}
\label{sec:metrics}

Since our target users are tech-savvy freelance translators and smaller language service providers (LSPs) who have only modest resources and little or no programming skills, we used MATEO \citep{vanroy2023}---a free, user-friendly web application---to generate the BLEU \citep{papineni_bleu_2001}, chrF \citep{popovic_chrf_2015}, TER \citep{snover_study_2006}, and COMET \citep{rei_comet_2020,rei_comet-22_2022} scores for all 12 translation outputs (9 local LLMs $+$ 3 baselines) for each of our language pairs. MATEO accepts parallel text files (source, reference, and up to four system outputs) and computes multiple automatic metrics through an intuitive web interface, making it an ideal tool for our target audience.

The  evaluation results are reported in Section ~\ref{sec:results} below and Appendix~\ref{app:evaluation}, as well as in the companion materials in our repository. We note that most of the pairwise score differences are statistically significant at the $p < 0.05$ level, lending confidence to the system rankings we derive.

We observe strong pairwise Pearson correlation between COMET and two string-based metrics, BLEU and chrF2, across all available systems for EN$\to$DE and EN$\to$RU: $r = 0.73$--$0.92$; and moderate correlation for EN$\to$JA and EN$\to$ZH: $r = 0.56$--$0.68$.\ The weakest correlation, for EN$\to$ZH, is likely attributable to the character-level properties of Chinese writing, which can introduce discrepancies between string-matching and neural evaluation approaches. Following common practice in MT evaluation, we set the string metrics aside at this point and used document- and sentence-level COMET scores in all subsequent evaluations.

\subsection{Summary of Document-Level Results}
\label{sec:results}

Table~\ref{tab:comet-main} presents document-level COMET scores for all the available translation outputs, including two different temperature settings for three ``reasoning models.'' The colored bars in Figure \ref{fig:four-comet-charts} display the best results for each of the 9 local LLMS set against the available baselines (gray bars).

\begin{table*}[h]
\centering
\small
\begin{tabular}{llcccc}
\toprule
\textbf{MT/LLM category} & \textbf{System} & \textbf{EN--DE} & \textbf{EN--RU} & \textbf{EN--JA} & \textbf{EN--ZH} \\
\midrule
\multirow{5}{*}{NMT: Frontier}
 & deepl-auto        & \textbf{90.07} & \textbf{90.72} & n/a & 89.31 \\
 & deepl-polite        & n/a & n/a & \textbf{91.01} & n/a \\
 & baidu-gen          & n/a   & n/a   & n/a   & \textbf{91.04} \\
 & baidu-pe          & n/a   & n/a   & n/a   & 90.95 \\
 & googletranslate    & 89.51 & n/a   & n/a   & n/a   \\
\midrule
LLM: Frontier
 & gpt-5.2            & 88.75 & 89.91 & 90.11 & 89.53 \\
\midrule
\multirow{9}{*}{LLM: Local}
 & aya-expanse-32b              & 88.18 & 89.14 & 89.84 & 89.19 \\
 & deepseek-r1-32b-t0.0         & 85.61 & 86.15 & 87.77 & 89.29 \\
 & deepseek-r1-32b-t0.8         & 84.30 & 84.63 & 86.89 & 88.67 \\
 & gemma3-27b                   & 88.05 & 89.13 & 89.67 & 89.34 \\
 & gpt-oss-20b-t0.0             & 86.90 & 86.23	& 89.00	& 88.85 \\
 & gpt-oss-20b-t1.0             & 88.01 & 88.45	& 89.27	& 88.62 \\
 & llama3.1-8b-instruct-fp16    & 86.23 & 87.01 & 87.36 & 87.11 \\
 & mistral-small3.2-24b         & \textbf{88.91} & 89.08 & 89.63 & 88.96 \\
 & qwen2.5-32b                  & 85.75 & 86.68 & 88.60 & 89.48 \\
 
 & qwen3-32b-t0.0               & 87.23 & 88.30 & 88.98 & 89.55 \\
 & qwen3-32b-t0.6               & 87.31 & 88.07 & 89.02 & 89.36 \\

 & translategemma-27b           & 88.31 & \textbf{89.55} & \textbf{90.34} & \textbf{89.88} \\
\midrule
\multirow{4}{*}{NMT: Local}
 & promt-23                       & 87.94 & \textbf{88.87} & \textbf{85.33} & \textbf{86.51} \\
 & neuraldesktop.2.1.0-auto             & n/a   & 88.26 & n/a   & n/a   \\
 & neuraldesktop.2.1.0-medical          & \textbf{88.68} & 88.66 & n/a   & n/a   \\
 & opus-cat\_2020-02-11           & 86.28 & 83.87 & n/a   & n/a   \\
 & opus-cat\_bt-2021-04-14           & n/a & 84.98 & n/a   & n/a   \\

\bottomrule
\end{tabular}
\caption{COMET-22 scores for all translation outputs. Highest values in each category are boldfaced.}
\label{tab:comet-main}
\end{table*}


Based on these scores, the best-performing local LLM for EN$\to$\{RU, ZH, JA\} is translategemma-27b, while mistral-small3.2-24b demonstrates the best performance for EN$\to$DE. We note that these local LLMs are ahead of GPT-5.2 for EN$\to$\{DE, ZH, JA\} and slightly behind it for EN$\to$RU. However, all the LLM outputs, including our local models and GPT-5.2, score significantly lower than the best commercial NMT systems: DeepL for EN$\to$\{DE, RU, JA\} and Baidu for EN$\to$ZH.

Classical scaling laws predict that model performance improves as a smooth function of parameter count \citep{kaplan_scaling_2020,hoffmann_training_2022}. Our document-level COMET scores (Table~\ref{tab:comet-main} and Figure \ref{fig:four-comet-charts}) offer only partial support for this prediction in the translation domain. The clearest scaling signal appears at the lower end of the parameter range: llama3.1-8b-instruct-fp16 is one of the weakest local LLM across all four language directions, with COMET scores 1.5–2.5 points below the best-performing models. This suggests that 8B parameters, even at full fp16 precision, remain insufficient for consistently competitive medical-domain translation.

Beyond this threshold, however, scaling behavior becomes decidedly non-monotonic. Among models in the 20B–32B range, parameter count alone is a poor predictor of translation quality. The 20B gpt-oss matches or exceeds several 32B models for EN→DE and EN→JA, while the translation-specialized translategemma-27b outperforms all 32B general-purpose models for EN→RU, EN→JA, and EN→ZH. Conversely, two 32B models, deepseek-r1 and qwen2.5, rank near the bottom for EN→DE and EN→RU despite their size advantage. These reversals suggest that at the parameter scales we tested, multilingual training data composition and task-specific optimization exert a stronger influence on translation quality than raw model size.

A language-specific pattern is also visible. Performance variance across models is wider for EN→DE and EN→RU than for EN→ZH, where the scores cluster more tightly.\ This may reflect asymmetries in the multilingual training corpora of different model families.\ Notably, the relatively strong EN→ZH performance of Qwen and DeepSeek models, both originating from Chinese organizations, hints at training-data effects that interact with scaling in complex ways.

These results caution against extrapolating general scaling laws to translation: model size alone is an unreliable proxy for translation quality, and empirical benchmarking on relevant language pairs remains essential.

We also tested three locally installable NMT systems used by some translators for offline workflows—Promt, OPUS-CAT, and NeuralDesktop—on the supported language pairs (Table~\ref{tab:comet-main}). Their document-level COMET scores vary widely. This variability may be due to the uneven quality of the base NMT models they use for different language directions \citep{tiedemann_opus-mt_2020} as well as the varying degrees of multilingual coverage and domain support.

We emphasize that all outputs reported here come from base models with no domain adaptation. Many systems, including commercial engines, local NMT, and local LLMs, can be fine-tuned or adapted to terminology and style constraints. Understanding adaptation regimes with modest resources is a natural next step and is the focus of ongoing research \citep{moslem_adaptive_2023,moslem_language_2024,vieira_how_2024,zheng_fine-tuning_2024,rios_instruction-tuned_2025}, with obvious practical implications. We plan to investigate various adaptation regimes, using our corpus, in the future.

We also acknowledge a limitation of our context-free, sentence-by-sentence evaluation approach: both the automatic and manual evaluation scores we report reflect segment-level quality and may not fully capture document-level coherence, consistency, or discourse-related translation phenomena \citep{castilho_online_2023,hu_source-primed_2025}. This is a deliberate constraint: many MT/LLM pipelines still operate sentence by sentence, and local inference costs make long-context prompting expensive and slow. At the same time, lack of discourse context can hurt pronoun resolution, lexical consistency, and terminology coherence. We discuss these limitations and plans for future work in Section~\ref{sec:discussion} below.

\section{Discussion, Limitations, and Future Work}
\label{sec:discussion}

Human evaluation has long been regarded as the gold standard in machine translation research. A companion paper to the present study therefore places human evaluation at the center of its analysis.

\paragraph{Human evaluation in the sequel.} A large-scale, blind, randomized human evaluation of all twelve outputs (nine local LLMs plus three baselines) plus reference translations on 100+ strategically selected sentences for EN→{DE, RU, ZH}, using a novel two-step direct-assessment protocol and a purpose-built Streamlit interface \citep{vanhorn2026}, is the focus of a companion paper currently in preparation. Two preliminary findings are worth flagging here: (i) while COMET is a strong system-level proxy for human judgment, it is a much weaker one at the sentence level, which is consistent with other recent findings; (ii) a non-trivial number of reference translations were judged by human experts to be inferior to the best system outputs. This has potential implications for reference-free quality estimation methods and related approaches \citep{lavie_findings_2025}, which could be particularly valuable for freelancers and smaller LSPs who may not always have access to high-quality reference translations.

\paragraph{Prompt and configuration effects as a workflow variable.} Unlike conventional NMT engines, local LLM MT makes prompt design part of the system. Our pilot experiments illustrate that prompt choices can materially shift quality and error profiles, and that ``native'' prompts suggested by a model do not necessarily optimize that model's outputs. For small teams, this implies that prompts should be versioned and treated like configuration code: changes should be evaluated on a fixed internal benchmark set.

\paragraph{Decoding temperature.} For reasoning models, temperature settings appear to be a double-edged sword: We find that $T=0.0$ is associated with a substantial number of translation-call failures in reasoning models.\ Changing the temperature to the model card or inference engine default ($T=0.6-1.0$) reduces the number of failed calls, but also appears to degrade the overall quality of the translation outputs.

\paragraph{When local LLM translation is viable.} Translation with local LLMs is most attractive when confidentiality constraints disallow cloud services, when marginal API costs are prohibitive, or when translators need predictable offline availability. Our results indicate that the best open models can provide high-quality drafts. However, for high-risk deliverables, such drafts should be used within a workflow that includes systematic validation rather than as a fully autonomous system.

\paragraph{Toward a fuller panel.} The selection of nine local LLMs reflected a snapshot of late-2025 availability; specialized recent releases such as SalamandraTA and Tower+ are obvious targets for the future.

\section{Concluding Remarks}
\label{sec:conclusion}

Using a real-world medical-domain corpus expanded to four target languages, we compared locally runnable open models against commercial MT, a frontier LLM baseline, and locally installable NMT systems.

Two conclusions stand out. First, translation-specialized open models and those trained on balanced multilingual data substantially  narrow the translation quality gap between local models and commercial engines, making translation with local LLMs increasingly viable under confidentiality and cost constraints. Second, prompt and decoding temperature choices can materially affect outcomes, so reproducibility and re-validation are essential.

\section*{Author Contributions}

YB conceived the study, led the corpus expansion and curation, conducted automatic evaluation of all the outputs with MATEO, and drafted the manuscript. RVH performed the German reference translation curation and contributed to the research strategy and evaluation design. MX performed the Chinese reference translation curation and contributed linguistic observations. AD implemented all local LLM inference operations, conducted prompt selection experiments, managed the technical infrastructure, and authored the description of these processes in Sections~\ref{sec:design} and~\ref{sec:generation}. All authors reviewed and approved the final manuscript.

\section*{Acknowledgments}

YB’s work was supported by NSF Grant No. SES-2336713. We reiterate our thanks to the Christopher \& Dana Reeve Foundation  for permission to use their linguistic resources in our experiments. We are indebted to Julian Hennemann for reviewing the EN–DE outputs and contributing valuable linguistic observations. We thank Chris Schertler for sharing the TMX and XLIFF documents for the German translation of PRG International. We are grateful to Sheila Castilho and to the participants in the graduate seminar on translation technologies taught at UGA in Fall 2025 for their input and advice. Finally, we sincerely thank the reviewers for their comments, most of which we have incorporated into this revised version.

\bibliography{Reeve-Mutilingual}
\bibliographystyle{eamt26}

\appendix

\section{The Reeve Foundation Multilingual Corpus}
\label{app:corpus}

Our corpus comprises 3,500 English source sentences aligned with their original professional translations into Russian and Japanese, as well as partially curated reference translations into German and Chinese. With the client's permission, we \href{https://github.com/YuriBalashov/reeve-mftc/tree/main/en-de-ru-ja-zh-reference}{\texttt{\textbf{release}}} it for noncommercial/academic use.


\section{Local Inference Setup}
\label{app:inference}

All local LLM experiments were conducted on a workstation running Ubuntu 24.04.4 LTS, equipped with NVIDIA GeForce RTX 3090 GPUs (24~GB VRAM), NVIDIA driver version 580.126.09, and CUDA 13.0. Each translation run used a single GPU. The inference engine was Ollama v0.14.2.

Table~\ref{tab:quantization} reports the quantization settings for each model. Models at or above 24B parameters were served in Ollama's default 4-bit quantization (Q4\_K\_M). The two smaller models, gpt-oss-20b and llama3.1-8b-instruct, were run at their default (unquantized) precision, as both fit within the 24~GB VRAM budget without compression. Llama3.1-8b-instruct was run at full fp16.

\begin{table}[ht]
\centering
\small
\begin{tabular}{lrl}
\toprule
\textbf{Model} & \textbf{Params} & \textbf{Quantization} \\
\midrule
aya-expanse-32b        & 32B & Q4\_K\_M \\
deepseek-r1-32b        & 32B & Q4\_K\_M \\
gemma3-27b             & 27B & Q4\_K\_M \\
gpt-oss-20b            & 20B & Default (unquant.)\\
llama3.1-8b-instruct   &  8B & fp16 \\
mistral-small3.2-24b   & 24B & Q4\_K\_M \\
qwen2.5-32b            & 32B & Q4\_K\_M \\
qwen3-32b              & 32B & Q4\_K\_M \\
translategemma-27b     & 27B & Q4\_K\_M \\
\bottomrule
\end{tabular}
\caption{Model quantization settings.}
\label{tab:quantization}
\end{table}

Table~\ref{tab:wallclock} reports wall-clock times for translating the full 1,143-sentence corpus per language direction on a single RTX 3090. Non-reasoning models completed each direction in approximately 19 to 40 minutes. The three reasoning-oriented models were substantially slower: qwen3-32b required 4.4 to 5.5 hours per direction, deepseek-r1-32b approximately 2.9 to 3.8 hours, and gpt-oss-20b approximately 43 to 64 minutes. This overhead reflects the models' internal multi-step processing, which generates extended reasoning traces before producing the final translation.

\begin{table}[ht]
\centering
\small
\begin{tabular}{lcccc}
\toprule
\textbf{Model} & \textbf{DE} & \textbf{RU} & \textbf{JA} & \textbf{ZH} \\
\midrule
aya-expanse     & 28m  & 27m  & 26m  & 23m  \\
deepseek-r1     & 3h47 & 3h44 & 3h28 & 2h51 \\
gemma3          & 27m  & 28m  & 25m  & 23m  \\
gpt-oss         & 58m  & 64m  & 61m  & 43m  \\
llama3.1-8b     & 23m  & 22m  & 22m  & 19m  \\
mistral-small   & 19m  & 22m  & 23m  & 20m  \\
qwen2.5         & 31m  & 39m  & 29m  & 19m  \\
qwen3           & 5h32 & 5h22 & 5h29 & 4h21 \\
translategemma  & 28m  & 29m  & 26m  & 24m  \\
\bottomrule
\end{tabular}
\caption{Wall-clock time per language direction (1,143 sentences, single RTX 3090).}
\label{tab:wallclock}
\end{table}

\section{Prompt Portfolio}
\label{app:prompts}

This appendix lists the full text of all candidate prompts used in the prompt selection pilot (Section~\ref{sec:prompt-selection}). Prompts p1--p4 and p6--p9 were elicited by meta-prompting the indicated model to generate its own preferred translation prompt. Prompts p10 and p11 are ``standard'' prompts from \citet{balashov_translation_2025}, in German and English respectively. Prompt p5 was generated by gemma3-27b and was selected as the best-performing prompt on average across all models (Section~\ref{sec:prompt-selection} and Table~\ref{tab:prompt-pilot}). Prompt p0 is the developer-recommended prompt for TranslateGemma \citep{finkelstein_translategemma_2026} and uses a different template format with explicit language code placeholders.

\begin{table*}[ht]
\centering
\small
\setlength{\tabcolsep}{3.5pt}
\begin{tabular}{lcccccccc}
\toprule
 & \textbf{aya} & \textbf{ds-r1} & \textbf{gemma3} & \textbf{gpt-oss} & \textbf{llama3.1} & \textbf{mistral} & \textbf{qwen2.5} & \textbf{qwen3}\\
\midrule
p1  & 86.30 & 83.01 & 86.93 & 85.71 & 83.27 & 85.96 & 83.10 & 85.19 \\
p2  & \textbf{86.55} & 83.32 & 86.94 & 86.48 & 83.00 & 86.03 & 84.66 & 85.73 \\
p3  & 85.78 & 83.17 & \textbf{87.05} & 86.24 & 83.76 & 86.31 & 84.19 & 85.36 \\
p4  & 85.75 & 83.00 & 86.75 & 85.86 & 83.90 & 86.41 & 84.39 & 84.86 \\
p5  & 85.87 & 83.86 & 86.75 & 86.50 & \textbf{84.56} & 86.56 & 84.16 & 85.53 \\
p6  & 85.90 & 82.77 & 86.32 & \textbf{86.91} & 83.89 & \textbf{86.63} & \textbf{84.84} & \textbf{85.74} \\
p7  & 86.00 & 83.43 & 85.65 & 86.42 & 84.10 & 86.42 & 84.28 & 85.34 \\
p8  & 85.97 & 82.70 & 86.69 & 86.14 & 83.96 & 86.51 & 83.82 & 84.97 \\
p9  & 86.06 & 83.25 & 85.29 & 86.49 & 83.96 & 86.04 & 83.79 & 85.43 \\
p10 & 85.64 & \textbf{83.99} & 86.22 & 86.54 & 83.45 & 86.49 & 84.27 & 84.81 \\
p11 & 86.47 & 83.72 & 86.51 & 86.38 & 83.23 & 86.26 & 83.70 & 85.00 \\

\bottomrule
\end{tabular}
\caption{COMET-22 scores for pilot experiments with 118 held-out test sentences (EN$\to$DE). The highest score for each model is boldfaced; the heat map of $z$-scored COMET improvements over the prompt mean across models is available in the companion materials.}

\label{tab:prompt-pilot}
\end{table*}

All prompts shown below are in their EN$\to$DE pilot form. For the main experiments, p5 was adapted to each target language by substituting the appropriate language name.

\medskip
\noindent\textbf{p0} (TranslateGemma developer-recommended): You are a professional \{SOURCE\_LANG\} (\{SOURCE\_CODE\}) to \{TARGET\_LANG\} (\{TARGET\_CODE\}) translator. Your goal is to accurately convey the meaning and nuances of the original \{SOURCE\_LANG\} text while adhering to \{TARGET\_LANG\} grammar, vocabulary, and cultural sensitivities. Produce only the \{TARGET\_LANG\} translation, without any additional explanations or commentary. Please translate the following \{SOURCE\_LANG\} text into \{TARGET\_LANG\}: \textbackslash n \textbackslash n \textbackslash n \{TEXT\}

\medskip
\noindent\textbf{p1} (Claude Opus 4.5): Translate the following English sentence to German. Output only the translation, nothing else.

\medskip
\noindent\textbf{p2} (GPT-5.2): You are a professional translator. Task: Translate the following sentence from English to German. Requirements: Output ONLY the German translation (no explanations, no quotes, no alternatives). Preserve meaning exactly; do not add or omit information. Keep names, numbers, units, and formatting unchanged. Use natural, idiomatic German. Use neutral formal register (Sie) unless the English clearly implies informal (du). Keep the same tense and voice; preserve emphasis if possible.

\medskip
\noindent\textbf{p3} (Gemini Pro 3): You are a professional translator proficient in English and German. Your task is to translate the provided English text into German. Constraints: Maintain the original tone and meaning. Output only the German translation. Do not include preambles, explanations, or enclosing quotes.

\medskip
\noindent\textbf{p4} (deepseek-r1-32b): You are a professional translator specializing in academic texts. Translate the following English text into German, maintaining its formal tone and precise terminology. Provide only the translation without any explanations, annotations, or transliterations.

\medskip
\noindent\textbf{p5} (gemma3-27b, selected for main experiments): Translate the following \{SOURCE\_LANG\} text into \{TARGET\_LANG\}. The target audience is academic/expert. Provide only the \{TARGET\_LANG\} translation, without any explanations, annotations, or transliterations.

\medskip
\noindent\textbf{p6} (gpt-oss-20b): You are a professional translator. Translate the following English text into German for an expert/academic audience. Output only the translated text, with no explanations, annotations, or transliterations.

\medskip
\noindent\textbf{p7} (mistral-small3.2-24b): Translate the following English text into German, maintaining an expert/academic level of language and terminology. Provide only the translation, without any explanations, annotations, or transliterations.

\medskip
\noindent\textbf{p8} (qwen2.5-32b): Translate the following English text into German for an expert/academic audience. Provide only the translation without any explanations, annotations, or transliterations.

\medskip
\noindent\textbf{p9} (qwen3-32b): Translate the following English text into German. The translation should be formal, precise, and tailored for an expert/academic audience. Provide only the translated text, with no additional explanations, annotations, or transliterations. Ensure technical terminology and nuanced meanings are accurately preserved.

\medskip
\noindent\textbf{p10} (standard prompt, German, from \citet{balashov_translation_2025}): Sie sind ein erfahrener \"Ubersetzer und \"ubersetzen f\"ur ein Fachpublikum. Bitte f\"ugen Sie Ihrer \"Ubersetzung keine Anmerkungen, Erkl\"arungen oder Transliterationen hinzu. Bitte \"ubersetzen Sie den folgenden Satz ins Deutsche:

\medskip
\noindent\textbf{p11} (standard prompt, English, from \citet{balashov_translation_2025}): You are an expert translator, translating for an expert audience. Please do not provide any annotations, explanations or transliterations in your translation. Please translate the following English sentence to German:

\medskip

Table~\ref{tab:prompt-pilot} presents the COMET scores for the pilot experiments with 118 held-out test sentences (EN$\to$DE).

\section{Levenshtein Ratio Matrices for Different Temperature Settings}
\label{app:temperature}

To assess the effect of decoding temperature on output consistency, we ran a sweep across 21 temperature values from $T=0.0$ to $T=1.0$ in increments of 0.05, using the 118-sentence EN$\to$DE pilot set (Section~\ref{sec:prompt-selection}). We generated one full translation of the pilot set at each temperature setting, then computed the mean pairwise Levenshtein ratio between all 21 x 21 combination of outputs. Each cell in the resulting matrix reports the mean Levenshtein ratio (computed over character sequences) across all 118 sentences for the corresponding temperature pair.

\begin{figure*} [h]
    \centering
    \includegraphics[width=\linewidth]{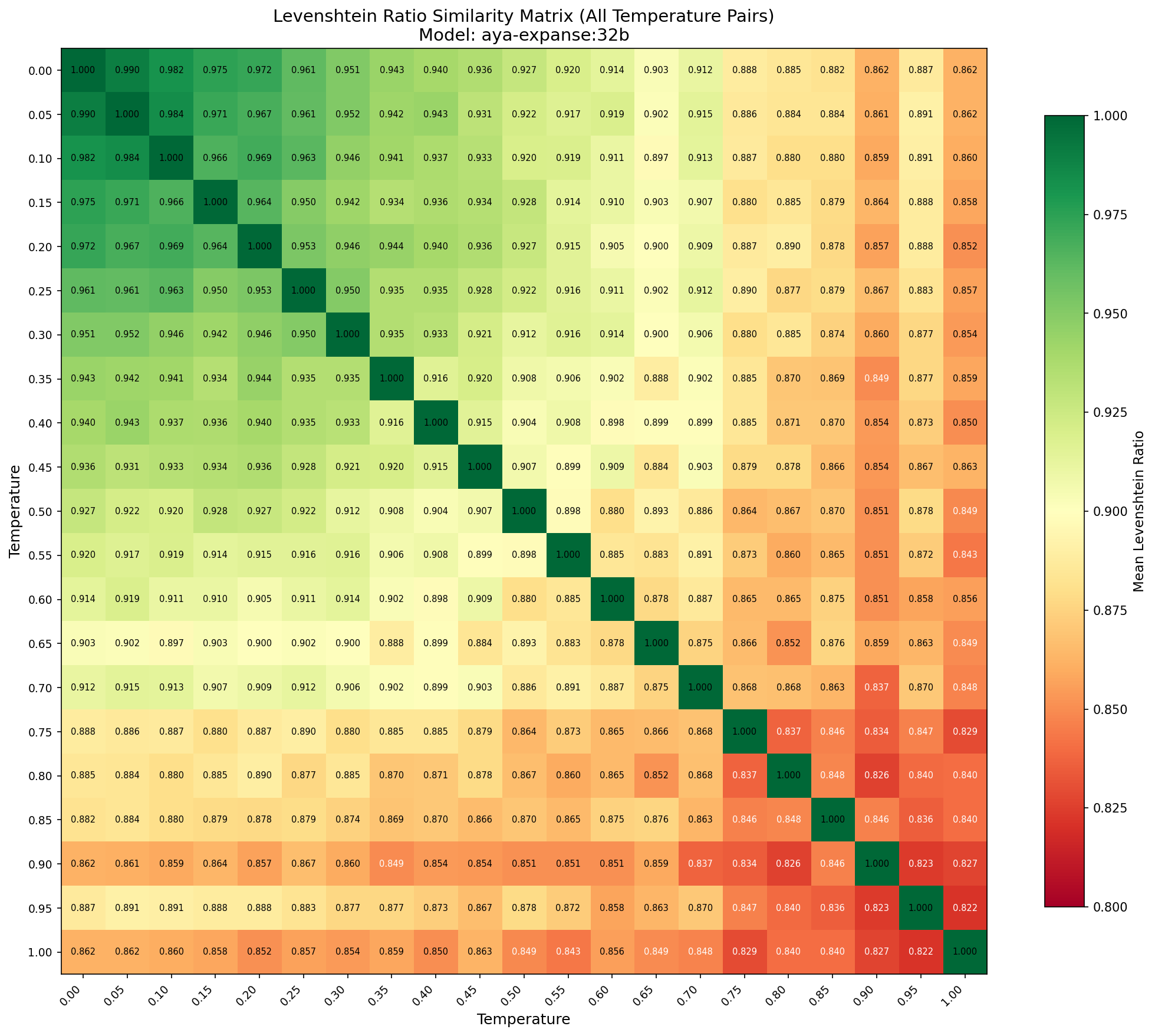}
    \caption{Levenshtein ratio similarity matrix for aya-expanse-32b across 21 temperature settings ($T=0.0$ to $T=1.0$, step 0.05), computed on the 118-sentence EN$\to$DE pilot set. Each cell reports the mean Levenshtein ratio (computed over character sequences) between outputs generated at the two corresponding temperatures. Values range from 0 (no overlap) to 1 (identical outputs); the diagonal is trivially 1.0. The $T=0.0$ row and column show the highest off-diagonal values, indicating that zero-temperature output is the centroid of the output distribution. Again these results suggest that temperature primarily affects output consistency rather than quality in translation; we therefore recommend users set temperature to 0.0.}
    \label{fig:levenshtein_ratio}
\end{figure*}

Figure~\ref{fig:levenshtein_ratio} shows the matrix for aya-expanse-32b. The row and column corresponding to $T=0.0$ show the highest similarity values throughout, confirming that zero-temperatures outputs are maximally close to outputs at all other temperatures. Similarity declines as the distance between compared temperatures increases, with the lowest values concentrated in the bottom-right corner of the matrix (high-temperature pairs).

\section{Translation Failures From Infinite ``Reasoning Loops''}
\label{app:failures}

Table~\ref{tab:failures} presents the number of failed translation calls resulting from prohibitively long ``thinking loops'' produced by three ``reasoning models'' on two temperature settings.

\begin{table*}[ht]
\centering
\small
\begin{tabular}{llccccc}
\toprule
\textbf{Model}  & \textbf{Temp} & \textbf{DE} & \textbf{RU} & \textbf{JA} & \textbf{ZH} & \textbf{Total} \\
\midrule
gpt-oss-20b     & 0.0 & 34  & 46  & 13  & 6  & 99  \\
                & 1.0 & 2 & 1 & - & - & 3\\

\midrule
deepseek-r1-32b & 0.0 & 1  & -  & -  & -  & 1\\
                & 0.8 & 1  & 3  & -  & -  & 4  \\
\midrule
qwen3-32b       & 0.0 & 6 & - & 3 & - & 9 \\
                & 0.6 & 4 & 3 & - & 1 & 8\\
\bottomrule
\end{tabular}
\caption{Number of failed translation calls due to ``thinking loops'' generated by three ``reasoning models.''}
\label{tab:failures}
\end{table*}

The scores reflected in Table~\ref{tab:comet-main} and Figure \ref{fig:four-comet-charts} reflect these failures: i.e. the models are penalized for the empty output lines. We think this is fair when reporting the actual performance of the models. To see how temperature settings affect the sentence-by-sentence quality of the outputs in cases of successful translation generation, we eliminated all outputs with at least one missing translation for each language pair and calculated the average document-level COMET score gains or losses denoted by `$T_0\to T_1$'. We also calculated the document-level average of the absolute values of such gains and losses: `$|T_0\to T_1|$'. The former gives a rough measure of the overall change in translation performance associated with the temperature change, while the latter represents the variation of such changes (Table~\ref{tab:gains-losses}).

\begin{table*}[ht]
\centering
\small
\begin{tabular}{llcccc}
\toprule
\textbf{Model}  & \textbf{Temp} & \textbf{DE} & \textbf{RU} & \textbf{JA} & \textbf{ZH} \\
\midrule
gpt-oss-20b     & $T_0\to T_{1.0}$ & $-0.32$  & $-0.14$  & $-0.39$  & $-0.54$\\
                & $|T_0\to T_{1.0}|$ & 1.82 & 2.16 & 1.87 & 2.17\\

\midrule
deepseek-r1-32b & $T_0\to T_{0.8}$ & $-1.26$  & $-1.46$  & $-0.87$  & $-0.61$\\
                & $|T_0\to T_{0.8}|$ & 4.35 & 5.42 & 3.83 & 2.45\\
\midrule
qwen3-32b       & $T_0\to T_{0.6}$ & $-0.00$  & $-0.10$  & $-0.09$  & $-0.13$\\
                & $|T_0\to T_{0.6}|$ & 2.29 & 2.26 & 2.16 & 1.89\\
\bottomrule
\end{tabular}
\caption{Document-level averages of COMET score gains or losses (`$T_0\to T_1$') and of their absolute values (`$|T_0\to T_1|$'} after excision of outputs with missing translations.
\label{tab:gains-losses}
\end{table*}

We emphasize that these numbers represent relative differences in the automatic scores and do not by themselves tell us anything about the actual quality of the outputs. In fact, the outputs from qwen3-32b and deepseek-r1-32b appears to be substandard for all language pairs except EN--ZH (Section~\ref{sec:results}).

\begin{figure*}[h]
\centering
\includegraphics[width=\linewidth]{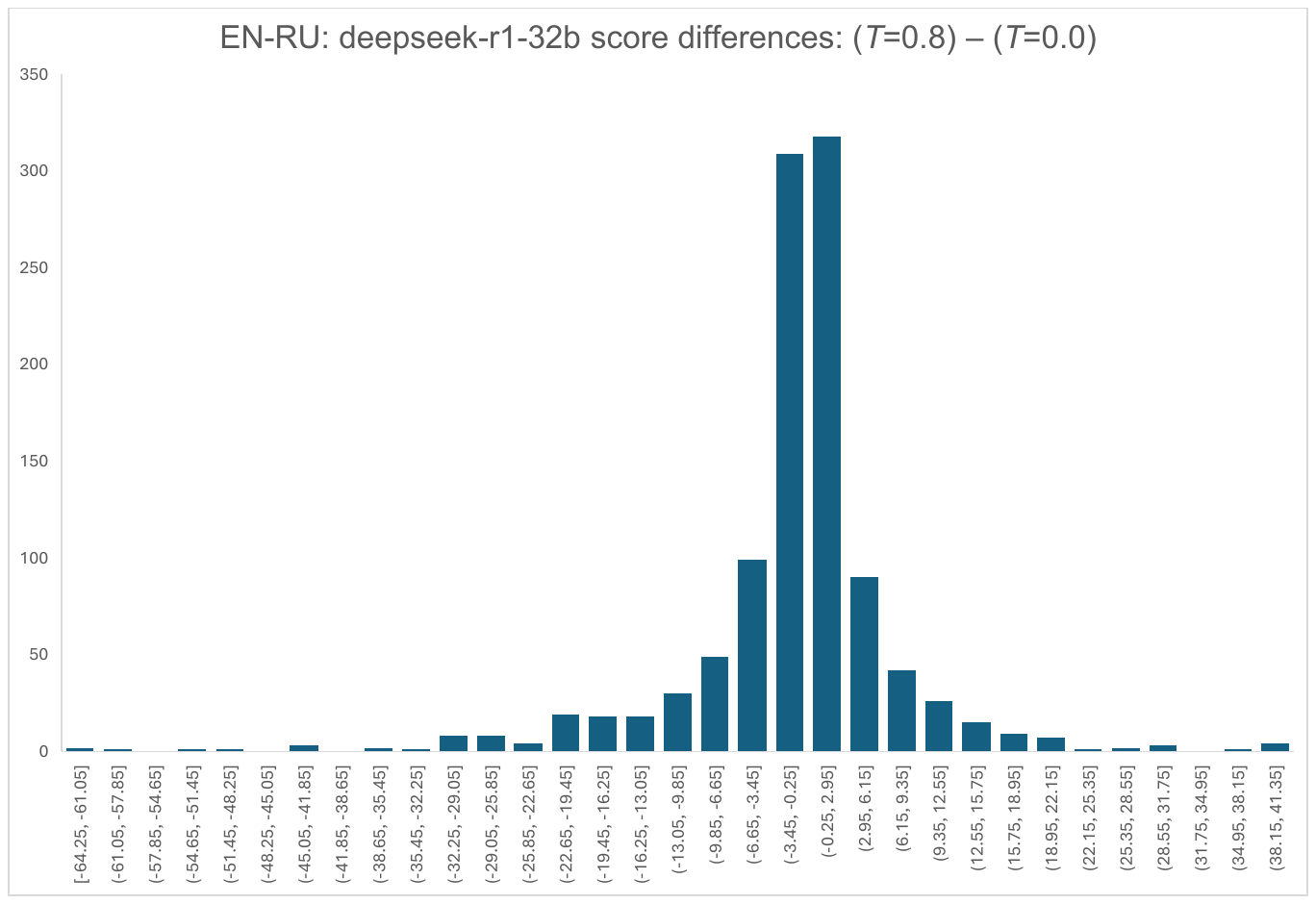}
\caption{Histogram of sentence-level COMET score differences between the EN--RU outputs from deepseek-r1-32b for $T=0.8$ and $T=0.0$. Three empty lines were removed.}
\label{fig:score-diff}
\end{figure*}

A typical histogram of the sentence-by-sentence COMET score differences representing a fine-grained breakdown of `$|T_0\to T_1|$' from Figure~\ref{fig:score-diff} is centered in the negative area thus indicating the overall loss of translation quality. We have reason to believe the long tails of the distributions of such differences may have contributed the most to the average losses reflected in Table~\ref{tab:gains-losses} and that these tails are associated with notable linguistic issues. We plan to review them in a  sequel to this paper which will include human evaluation of a substantial number of our translation outputs.

\section{Automatic Evaluation: Charts}
\label{app:evaluation}

\begin{figure*}[h]
\centering
\includegraphics[width=\linewidth]{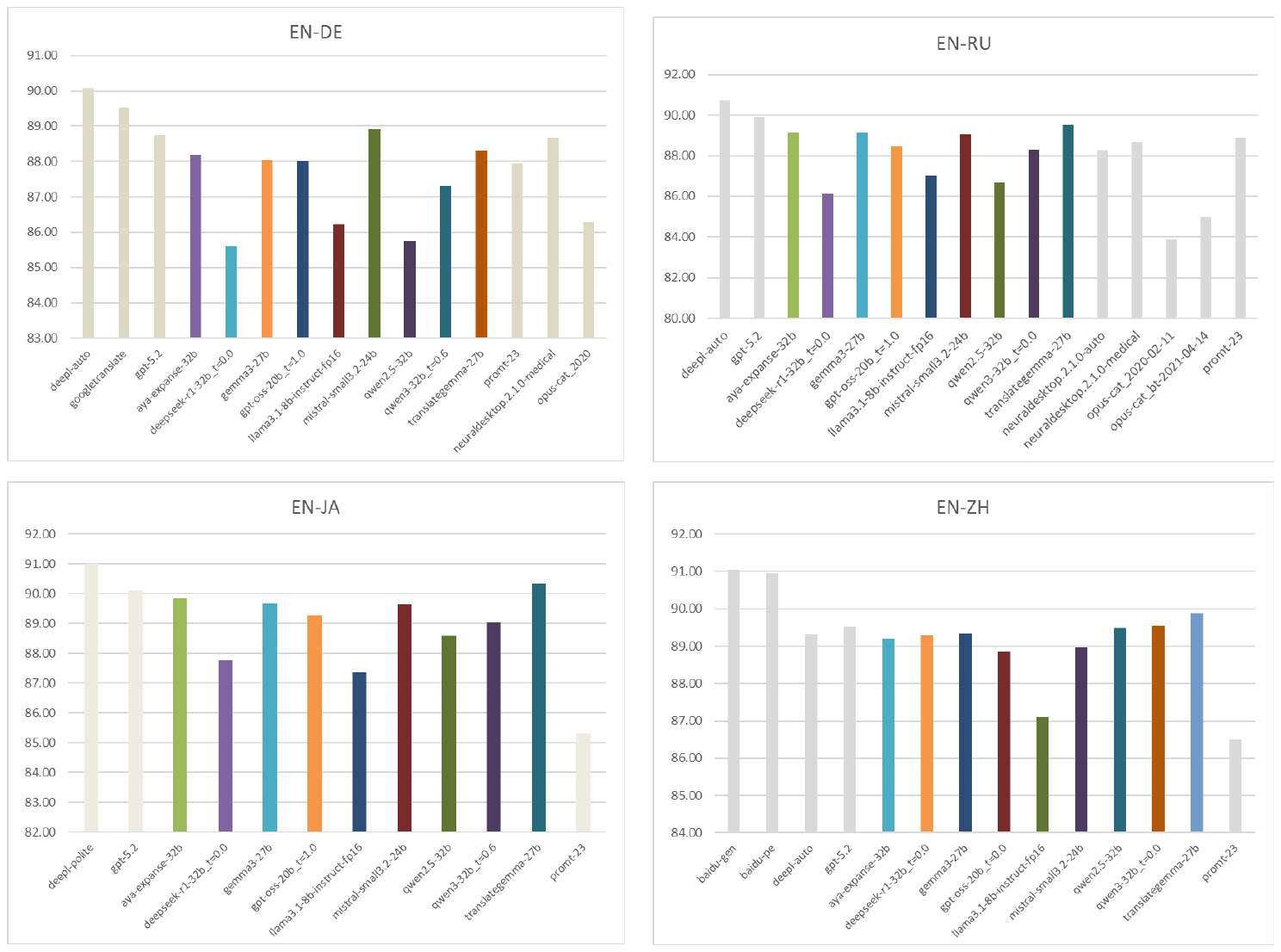}
\caption{COMET-22 scores for translation outputs. Colored bars: best outputs from each of the nine local LLMs. Gray bars: available baselines.}
\label{fig:four-comet-charts}
\end{figure*}

Figure \ref{fig:four-comet-charts} displays COMET-22 scores for translation outputs (from those listed in Table~\ref{tab:comet-main}). Colored bars represent the best outputs from each of the 9 local LLMs, gray bars the outputs from the baselines for each translation direction.

\end{document}